\documentclass[11pt]{article}
\usepackage[final]{acl}  % Using review option for line numbers during review process
\usepackage[colorinlistoftodos]{todonotes}
\usepackage{booktabs}
\usepackage[T1]{fontenc}
\usepackage[utf8]{inputenc}
\usepackage{array}
\usepackage{makecell}
\usepackage{tabularx}
\usepackage{adjustbox}
\usepackage{multirow} % Added for \multirow command

\title{A Culturally-Rich Romanian NLP Dataset from "Who Wants to Be a Millionaire?" Videos}

\author{\textbf{Alexandru-Gabriel Ganea} \textbf{Antonia-Adelina Popovici} \textbf{Adrian-Marius Dumitran} \\
      {University of Bucharest Faculty of Mathematics and Computer Science}}

\begin{document}
\maketitle

\begin{abstract}
Large Language Models (LLMs) demonstrate varying performance across languages and cultural contexts. This study introduces a novel, culturally-rich, multilingual dataset derived from video recordings of the Romanian game show "Who Wants to Be a Millionaire?" (Vrei să fii Milionar?).  We employed an innovative process combining optical character recognition (OCR), automated text extraction, and manual verification to collect question-answer pairs, enriching them with metadata including question domain (e.g., biology, history), cultural relevance (Romanian-specific vs. international), and difficulty. Benchmarking state-of-the-art LLMs, including Romanian-adapted models, on this dataset revealed significant performance disparities: models consistently achieve higher accuracy (80-95\%) on international questions compared to Romanian-specific cultural questions (50-75\%). We further investigate these differences through experiments involving machine translation of Romanian questions into English and cross-lingual tests using a comparable dataset in French. Our findings underscore the impact of cultural context and data source on LLM performance and offer practical insights for building robust, culturally-aware multilingual NLP systems, especially in educational domains. The dataset is publicly available at Hugging Face.

\end{abstract}

\section{Introduction}

The rapid advancement of large language models (LLMs) has transformed many NLP tasks, including question answering, summarization, and translation. However, most evaluations focus on high-resource languages like English, leaving a gap in understanding LLM performance in lower-resource and culturally diverse contexts.

To address this, we introduce a Romanian-language dataset derived from \textit{Vrei să fii milionar?}, the local version of \href{https://en.wikipedia.org/wiki/Who_Wants_to_Be_a_Millionaire}{"Who Wants to Be a Millionaire?"} \footnote{\url{https://en.wikipedia.org/wiki/Who_Wants_to_Be_a_Millionaire}}. Publicly available on \href{https://huggingface.co/datasets/WWTBM/wwtbm}{Hugging Face} \footnote{\url{https://huggingface.co/datasets/WWTBM/wwtbm}}, the dataset was compiled directly from video recordings, requiring structured data extraction from a dynamic visual format rather than standard text corpora. It reflects authentic language in a culturally specific, conversational quiz show setting. Each question is annotated for cultural relevance—Romanian-specific vs. international—and labeled by difficulty (easy, medium, hard).

Evaluating LLMs in Romanian poses unique challenges, including complex grammar, rich vocabulary, and limited NLP resources. This study investigates how well current models handle Romanian-language multiple-choice question answering (MCQA), with emphasis on cultural knowledge and the performance impact of Romanian-specific fine-tuning. We benchmark state-of-the-art open-source LLMs, including models adapted for Romanian, to explore the effects of language and cultural grounding.

Our findings offer insights into the strengths and limitations of multilingual LLMs in culturally rich settings and contribute toward building more inclusive, robust NLP systems for underrepresented languages like Romanian.

\section{Related Work}

\subsection{Linguistic and Cultural Diversity in LLM Evaluation}

While LLMs such as GPT-4 \cite{openai2023gpt4}, PaLM \cite{chowdhery2022palm}, and LLaMA \cite{touvron2023llama} have shown impressive performance across a wide range of benchmarks, most evaluation datasets are heavily skewed toward English or other high-resource languages \cite{wang2022super,nangia2021wt5}. This imbalance limits our understanding of how well these models generalize to linguistically and culturally diverse contexts. Several efforts have highlighted this issue and proposed multilingual benchmarks such as XTREME \cite{hu2020xtreme} and XGLUE \cite{liang2020xglue}, but these often lack deep cultural grounding and fail to include many underrepresented languages.

More recent work has turned attention to culturally situated evaluation. For example, the BOLD benchmark\cite{dhamala2021bold} introduces cultural dimensions to evaluate biases in language models.

\subsection{Romanian NLP Datasets and Resources}

Romanian, as a mid-resource language, has seen increasing support through dedicated datasets and benchmarks. The RoBERT model \cite{dumitrescu2020birth}, based on the original BERT architecture and trained specifically for Romanian, is not to be confused with RoBERTa \cite{liu2019roberta}, a robustly optimized BERT approach pretrained on larger, English-focused corpora. Foundational resources like CoRoLa \cite{tufis2017corola} have further enabled robust pretraining and evaluation. More recently, the "Vorbesti Românește?" initiative \cite{masala2024openllmrotechnicalreport} introduced a large-scale instruction-tuned Romanian benchmark suite, along with open-source models and datasets that significantly advance the capabilities of Romanian LLMs across multiple evaluation categories.

\subsection{Language Models for Romanian and Other Underrepresented Languages}

Several transformer-based models have been trained or adapted specifically for Romanian, such as RoBERT and DistilRoBERTa \cite{pitur2021romanian} include Romanian as part of their training data, although their performance on Romanian tasks varies greatly depending on the domain and task complexity.

In the broader context of low-resource or typologically diverse languages, projects like Masakhane \cite{nekoto2020participatory} and AmericasNLP \cite{mager2021findings} advocate for community-driven, culturally-aware NLP development. These projects underscore the importance of local expertise and participatory research in building fair and effective NLP tools for underrepresented languages.

In line with efforts to support underrepresented languages, our work emphasizes the value of non-traditional data sources for NLP development. Unlike most Romanian-language datasets, which are typically derived from written corpora, our approach leverages video recordings of a quiz show to extract question-answer pairs. This not only provides a more dynamic linguistic sample—complete with spoken language nuances and cultural references—but also demonstrates how alternative modalities can be harnessed to build more representative NLP resources.

\section{Dataset Creation}

We developed a multilingual dataset centered on the Romanian edition of the quiz show \href{https://huggingface.co/datasets/WWTBM/wwtbm}{\textit{Who Wants to Be a Millionaire?}}. The core of this resource is the highly curated and annotated Romanian dataset. To facilitate comparative and cross-lingual analysis, we also translated this primary Romanian data into English and curated parallel datasets in English and French, following a similar structure where possible. The creation process for the primary Romanian dataset involved multiple steps, outlined below.

\subsection{Data Collection and Frame Extraction}

The final dataset comprises 1000 multiple-choice questions collected from publicly available video recordings. Approximately 400 questions originated from episodes aired between 2011–2012 \footnote{\url{https://drive.google.com/drive/folders/1zjO0P_awwSm52uWQKc9gMpEh25LrVfnC}}, obtained via Google Drive, and around 600 questions were extracted from 2018–2019 episodes downloaded from YouTube. \footnote{\url{https://www.youtube.com/playlist?list=PLvC_Gs1fsycShkx65zNpIqUhlgTuVI8UB}}

\begin{figure}[ht]
    \centering
    \includegraphics[width=0.50\textwidth]{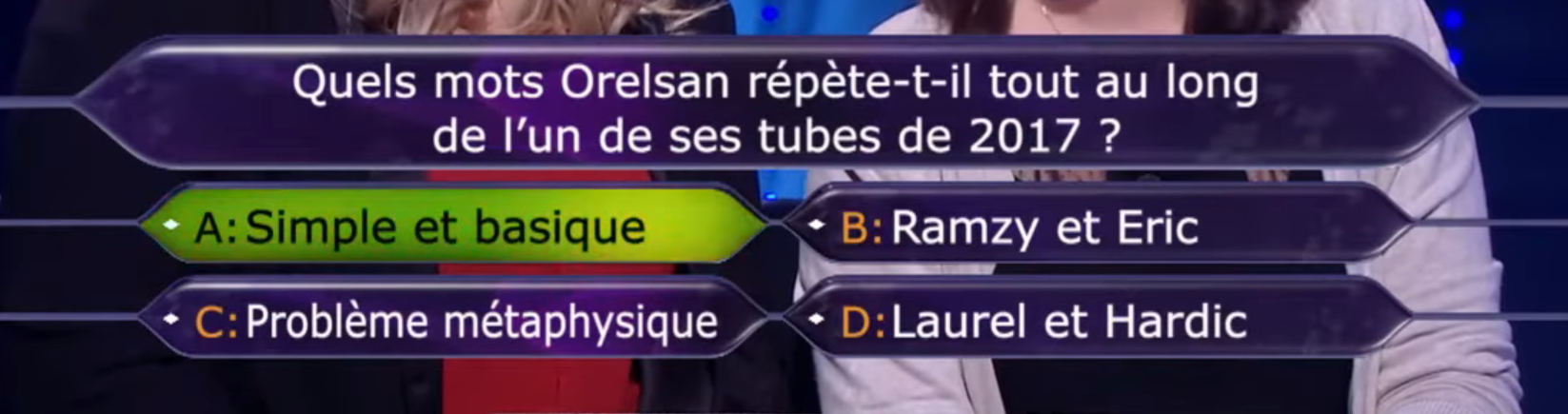}
    \caption{Example of question frame}
    \label{fig:example_question}
\end{figure}
Each video was analyzed frame-by-frame to capture screenshots precisely when the correct answer turned green, signifying correctness. To avoid redundancy, subsequent frames (approximately 500 frames per question) were skipped. Captured images were manually reviewed to ensure accuracy.

\subsection{Text Extraction and Diacritic Correction}

Text was extracted using Google's \texttt{gemini-1.5-flash-002} \citep{geminiteam2024gemini15unlockingmultimodal} and structured into Q\&A pairs. Romanian diacritics were automatically restored using a \href{https://huggingface.co/iliemihai/mt5-base-romanian-diacritics}{Romanian finetuned model} \texttt{mt5-base-romanian-diacritics} \footnote{\url{https://huggingface.co/iliemihai/mt5-base-romanian-diacritics}} version of MT5 \citep{xue2021mt5massivelymultilingualpretrained}, followed by minimal manual correction.

\subsection{Duplicate Removal}

Duplicates were identified by calculating cosine similarity using embeddings generated with \texttt{jina-embeddings-v3} \citep{sturua2024jinaembeddingsv3multilingualembeddingstask}. Questions exceeding a similarity threshold of 0.9 were manually reviewed, resulting in the removal of approximately 10 duplicates.

\subsection{Metadata Annotation}

Questions were enriched with relevant metadata, including:
\begin{itemize}
    \item \textbf{Episode Air Date}.
    \item \textbf{Monetary Value (RON)}: Used as a proxy for question difficulty.
    \item \textbf{Difficulty Level}: Easy, medium, and hard (based on monetary value).
    \item \textbf{Category}: Art and Culture, Cinematography, Gastronomy...
    \item \textbf{Cultural context}: Romanian or International. For example if a questions asks about "Titanic" ->International whereas if the question is about "Filantropica" -> then its Romanian.
\end{itemize}

\begin{figure}[ht]
    \centering
    \includegraphics[width=0.40\textwidth]{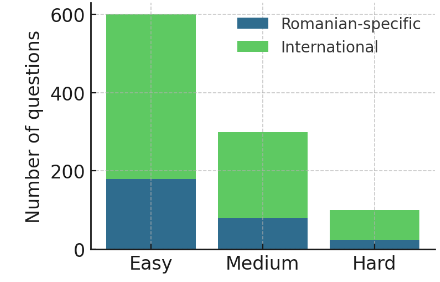}
    \caption{Distribution of difficulty levels across Romanian-specific and international cultural contexts. Note the limited number of "hard" questions reflecting contestant dropout at higher quiz levels.}
    \label{fig:difficulty_distribution}
\end{figure}

The distribution of questions across difficulty levels (Figure~\ref{fig:difficulty_distribution}) shows the distribution of questions across difficulty levels 

\subsection{Cultural Context Categorization}

Questions were automatically classified as Romanian-specific or international using \texttt{Qwen2.5-72B-Instruct} and manually validated, resulting in a 28.4\% / 71.6\% split.

\subsection{Topic Categorization}

\begin{figure}[ht]
    \centering
    \includegraphics[width=0.48\textwidth]{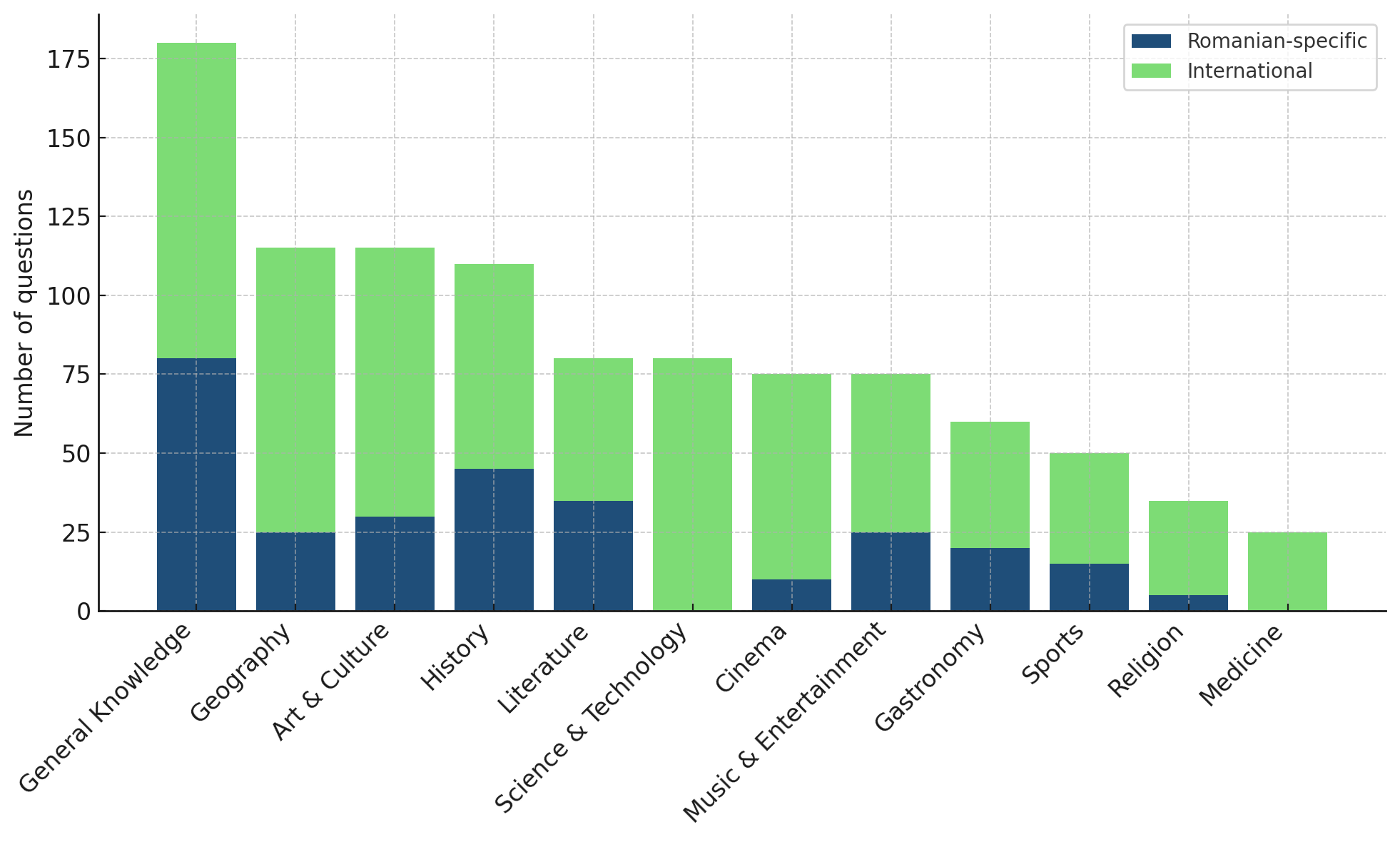}
    \caption{Distribution of questions across topic categories.}
    \label{fig:topic_distribution}
\end{figure}
Questions were automatically assigned to 12 topic domains using \texttt{Qwen2.5-72B-Instruct}, followed by manual verification and correction where necessary. The topic distribution is summarized in Figure~\ref{fig:topic_distribution}, which shows the number of Romanian-specific and international questions across each domain.

\section{Methodology}

We evaluated the performance of several state-of-the-art LLMs, specifically chosen to represent both general multilingual models and Romanian-specific adaptations. The evaluation framework, including model selection, configurations, prompting strategy, and testing conditions, is detailed below.

\subsection{LLM Selection and Configuration}

We benchmarked a diverse range of models (7B-72B parameters) covering various architectures, capabilities, and Romanian adaptations.

\begin{table}[ht]
\resizebox{\columnwidth}{!}{
\begin{tabular}{ll}
\toprule
\textbf{Category} & \textbf{Model} \\
\midrule
\multirow{8}{*}{General Multilingual} % Corrected multirow
  & Llama-3.1-8B-Instruct \citep{llama3_1herdmodels} \\
  & Llama-3.3-70B-Instruct \citep{llama3_3herdmodels} \\
  & Gemma2-9B-Instruct \citep{gemmateam2024gemma2improvingopen} \\
  & Mistral-7B-Instruct-v0.1 \citep{jiang2023mistral7b} \\
  & Aya-23-8B \citep{aryabumi2024aya23openweight} \\ % covers 23 languages
  & EuroLLM-9B-Instruct \citep{martins2024eurollmmultilinguallanguagemodels} \\ % focuses on European languages
  & Qwen2.5-72B-Instruct \citep{qwen2025qwen25technicalreport} \\
  & Phi-4 \citep{abdin2024phi4technicalreport} \\ % optimized for reasoning tasks
\midrule
\multirow{4}{*}{Romanian Fine-tuned} % Corrected multirow
  & RoGemma2-9B-Instruct \citep{masala2024openllmrotechnicalreport} \\
  & RoLlama3.1-8B-Instruct \citep{masala2024openllmrotechnicalreport} \\
  & RoMistral-7B-Instruct \citep{masala2024openllmrotechnicalreport} \\
  & Pansophic-1-preview \citep{pansophic2024preview} \\ % Romanian-optimized
\bottomrule
\end{tabular}}
\caption{Evaluated LLMs used in the benchmark.}
\label{tab:llm_evaluated_models}
\end{table}

% \textbf{Models Evaluated:}
% \begin{itemize}
%     \item \textit{General Multilingual Models:}
%         \begin{itemize}
%             \item \textbf{Llama Family:} Llama-3.1-8B-Instruct \citep{llama3_1herdmodels}, Llama-3.3-70B-Instruct \citep{llama3_3herdmodels}.
%             \item \textbf{Gemma2-9B-Instruct} \citep{gemmateam2024gemma2improvingopen}
%             \item \textbf{Mistral-7B-Instruct-v0.1} \citep{jiang2023mistral7b}
%             \item \textbf{Aya-23-8B} (covers 23 languages) \citep{aryabumi2024aya23openweight}
%             \item \textbf{EuroLLM-9B-Instruct} (emphasizes European languages) \citep{martins2024eurollmmultilinguallanguagemodels}
%             \item \textbf{Qwen2.5-72B-Instruct} \citep{qwen2025qwen25technicalreport}
%             \item \textbf{Phi-4} (focuses on reasoning tasks) \citep{abdin2024phi4technicalreport}
%         \end{itemize}
%     \item \textit{Romanian Fine-tuned Models:}
%         \begin{itemize}
%             \item \textbf{OpenLLM-Ro Adaptations:} RoGemma2-9B-Instruct, RoLlama3.1-8B-Instruct, RoMistral-7B-Instruct \citep{masala2024openllmrotechnicalreport}.
%             \item \textbf{Pansophic-1-preview} \citep{pansophic2024preview}.
%         \end{itemize}
% \end{itemize}

Inference for the largest models (\texttt{Qwen2.5-72B}, \texttt{Llama-3.3-70B}) utilized the Hyperbolic API\footnote{\url{https://hyperbolic.xyz}}; others were evaluated locally (Kaggle, 2x NVIDIA T4 GPUs).

\subsection{Experiments}

We evaluated model performance across multiple dimensions to assess capabilities in handling Romanian language and cultural nuances. The experiments were structured along three main axes:

\begin{enumerate}
    \item \textbf{Category-Based Evaluation:} Performance was measured separately for each of the 12 annotated topic categories (e.g., Art, History, Science) to identify domain-specific strengths and weaknesses.

    \item \textbf{Difficulty-Based Evaluation:} Models were assessed across easy, medium, and hard difficulty levels (derived from game show monetary value) to evaluate robustness to varying challenge levels.

    \item \textbf{Cultural Context Evaluation:} We performed separate evaluations on Romanian-specific versus international questions to isolate the impact of cultural context and identify potential cultural knowledge gaps.
\end{enumerate}

To further investigate cross-lingual and cultural generalization, we also conducted experiments using:
\begin{itemize}
    \item \textbf{Comparable French and English Datasets:} To observe if performance patterns generalized across other languages, including another Romance language.
    \item \textbf{Romanian-English Translation:} We translated the Romanian dataset into English and re-evaluated models to help disentangle linguistic understanding challenges from cultural knowledge factors.
\end{itemize}
\subsection{Prompt Design}

To ensure consistent and comparable results across all models, we standardized the prompt format. Each prompt included the question text followed by four multiple-choice answer options (\texttt{a, b, c, d}). Models were explicitly instructed to respond only with the letter corresponding to their chosen answer. The prompt was structured as follows:

\begin{quote}
\textit{Please respond with only the letter corresponding to the correct answer. Do not include any additional text, explanations, or punctuation.}
\end{quote}

Additionally, a system-level instruction was included to further guide the model behavior:

\begin{quote}
\textit{You are a master of answering multiple-choice questions who responds only with the letter corresponding to the correct answer.}
\end{quote}

Both prompts and system instructions were translated into Romanian to match the evaluation language of the dataset.
All evaluations were conducted using a zero-shot prompting strategy.

\subsection{Model Inference Configuration}

To maintain consistent inference behavior and minimize randomness, the following parameters were uniformly applied across all model runs:

\begin{itemize}
    \item \textbf{Max Tokens}: 1 (restricting output to one character).
    \item \textbf{Temperature}: 0 (fully deterministic responses).
 %   \item \textbf{Top-p}: 0.9 (allowing focused but slightly varied predictions).  
 %cu tempeature 0 top p nu stiu daca mai e relevant 
\end{itemize}

These settings ensured consistent, concise, and deterministic outputs, facilitating accurate performance comparisons.

\subsection{Evaluation Pipelines}

Evaluation procedures differed based on model access. \textbf{API-based models} (\texttt{Qwen2.5-72B}, \texttt{Llama-3.3-70B}) were queried via the Hyperbolic API using JSON requests, with built-in retries for rate limits. \textbf{Locally executed models} processed structured prompts directly using Hugging Face Transformers and local tokenizers. For both pipelines, failed or incomplete predictions were consistently marked with a placeholder (\texttt{'x'}) for standardized error handling.
\subsection{Performance Metrics}

Model predictions were systematically compared against ground-truth answers. Accuracy scores were calculated separately for each testing condition (cultural context, difficulty, and category), providing insights into models’ relative strengths and limitations, particularly concerning their understanding of Romanian language and cultural-specific knowledge.

\section{Results and Analysis}

We evaluate model performance comprehensively, focusing separately on cultural context, topic categories, and question difficulty.

\subsection{Performance by Cultural Context} \label{sec:cultural_context_results}
Table~\ref{tab:model_cultural_comparison} reveals model accuracy based on cultural context (Romanian-specific vs. international).

\begin{table}[ht]
\resizebox{\columnwidth}{!}{
\begin{tabular}{lccc}
\toprule
\textbf{Model} & \textbf{Romanian} & \textbf{International} & \textbf{Overall} \\
\midrule
RoGemma2-9B & 60.3 & 91.5 & 82.8 \\
Gemma2-9B & 62.8 & 89.2 & 81.8 \\
Llama-3.1-8B & 52.3 & 79.1 & 71.6 \\
RoLlama3.1-8B & 48.7 & 84.2 & 74.3 \\
Pansophic-1 & 50.5 & 79.8 & 71.6 \\
Aya-23-8B & 46.9 & 78.1 & 69.3 \\
Mistral-7B & 32.1 & 59.8 & 52.0 \\
RoMistral-7B & 49.1 & 84.1 & 74.3 \\
EuroLLM-9B & 70.7 & 88.2 & 83.3 \\
Qwen2.5-72B & 63.9 & 94.4 & 85.9 \\
Llama-3.3-70B & \textbf{75.8} & \textbf{96.5} & \textbf{90.7} \\
Phi-4 & 56.3 & 85.2 & 77.1 \\
\bottomrule
\end{tabular}}
\caption{Model accuracy (\%) by cultural context.}
\label{tab:model_cultural_comparison}
\end{table}

\paragraph{Analysis and Observations:}

\begin{itemize}
    % \item \textbf{Significant Cultural Gap:} All models perform substantially better on international questions (accuracies often 80-96\%) compared to Romanian-specific ones (50-75\% or lower), clearly demonstrating the challenge posed by cultural context.
    \item \textbf{Significant Cultural Gap:} We investigated whether there is a statistically significant difference in model performance between Romanian-specific and international questions.

    The null hypothesis states that there is no difference in mean accuracy between Romanian-specific and international questions across models. For each of the 12 models, we computed the mean accuracy separately on the Romanian-specific and international subsets, and then performed a paired t-test on these per-model differences.

    The test rejected the null hypothesis, showing a statistically significant difference in performance ($t(11) = 18.82$, $p < 0.001$). This indicates that models perform significantly better on international questions than on Romanian-specific ones, highlighting the challenge posed by cultural context.

    \item \textbf{Top Model Performance:} While \texttt{Llama-3.3-70B-Instruct} achieves the highest overall (90.7\%) and international (96.5\%) scores, its accuracy still drops significantly on Romanian-specific questions (75.8\%), indicating that large scale does not fully overcome the need for specific cultural knowledge.

\item \textbf{Difficulty Distribution Paradox:} Crucially, this performance gap exists \textit{despite} the Romanian-specific subset having proportionally fewer 'Hard' questions compared to the international set (as shown in Figure~\ref{fig:difficulty_distribution}). This strongly suggests that Romanian-specific questions, even those classified as 'Easy' or 'Medium' based on game progression, present an intrinsic challenge likely due to reliance on specific cultural facts, historical figures, or linguistic nuances less represented in models' training data.

    \item \textbf{Fine-tuning Effects:} Romanian fine-tuning yields mixed results. \texttt{RoGemma2-9B} and \texttt{RoMistral-7B} improved notably over their base models, mainly on international questions, suggesting better linguistic adaptation rather than direct cultural knowledge gains.

    \item \textbf{Smaller Model Struggles:} Smaller models like \texttt{Mistral-7B} and \texttt{Aya-23-8B} struggle significantly, particularly with Romanian-specific questions, highlighting the combined difficulty of handling a mid-resource language and specific cultural content.
\end{itemize}

These results underscore the importance of evaluating LLMs within specific cultural contexts.
Further experiments exploring the impact of fine-tuning in few-shot settings are detailed in Appendix~\ref{sec:appendix_few_shot}.

\subsection{Performance by Topic Categories}

\begin{figure}[ht]
  \centering
  % PDF keeps text crisp; PNG is fallback for draft mode
  \includegraphics[width=\columnwidth]{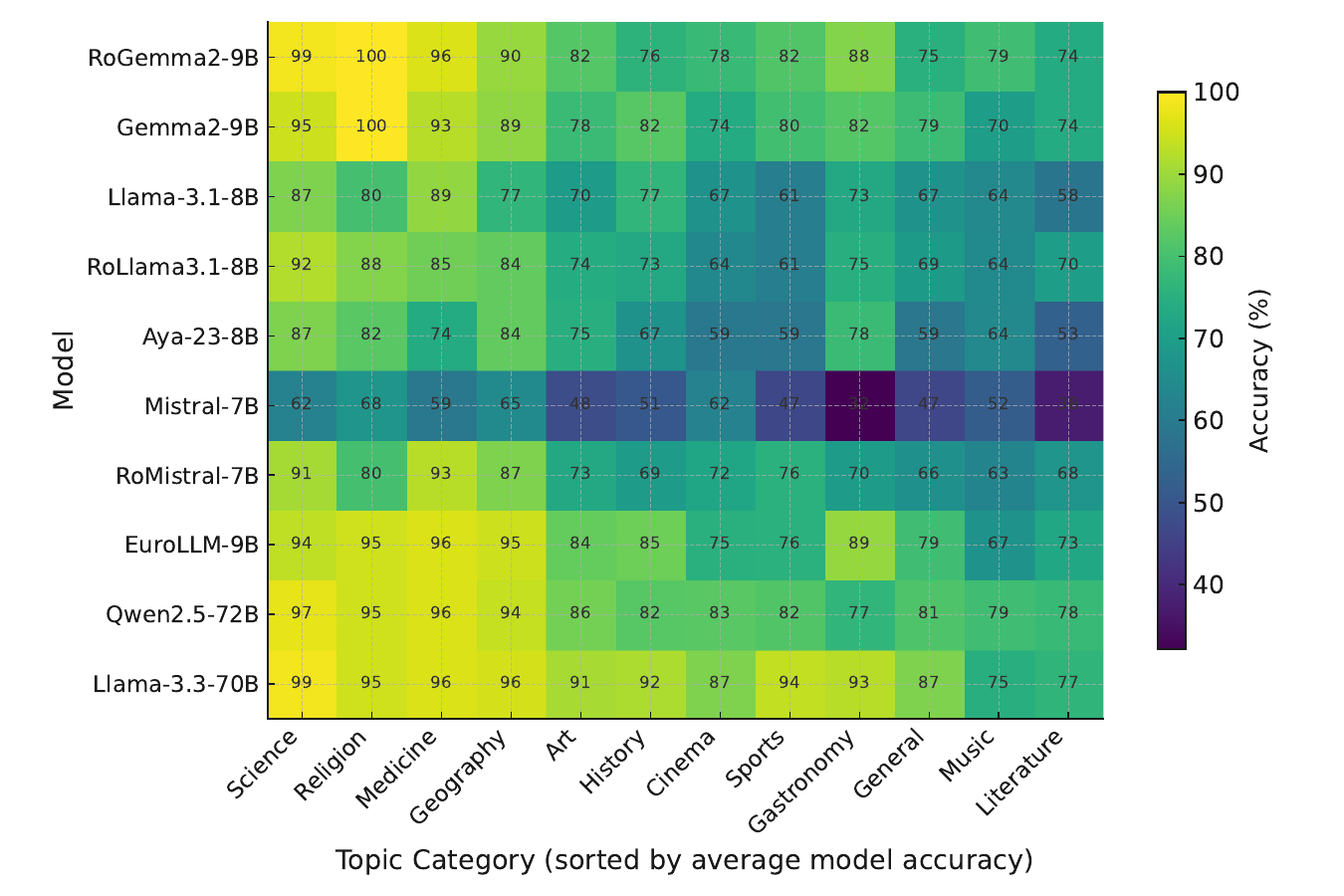}
  \caption{Model accuracies by topic (categories sorted by average score).}
  \label{fig:model_category_heatmap_sorted}
\end{figure}

To understand performance variations across different knowledge domains, we analyzed model accuracy for each of the 12 topic categories. Figure~\ref{fig:model_category_heatmap_sorted} presents these results as a heatmap, offering a visual comparison across models and topics. The topic categories on the x-axis are sorted by descending average accuracy across all models, indicating generally "easier" topics on the left and "harder" topics on the right.

\paragraph{Analysis and Observations:}

The heatmap highlights several key trends:

\begin{itemize}
    \item \textbf{Topic Difficulty Gradient:} A clear performance gradient exists, with topics like Science, Religion, and Medicine (left side) generally yielding higher accuracies than Literature, Music, and General Culture (right side).
    \item \textbf{Model Scale Matters:} Larger models (\texttt{Llama-3.3-70B}, \texttt{Qwen2.5-72B}) display consistently high performance (more yellow/bright green) across nearly all categories.
    \item \textbf{Specific Model Weaknesses:} Smaller models, particularly \texttt{Mistral-7B}, show significant weaknesses (darker cells) in multiple, often lower-performing, categories.
    \item \textbf{Fine-tuning Effects Visually:} Romanian fine-tuning shows noticeable benefits for \texttt{RoMistral-7B} compared to its base. Improvements for \texttt{RoGemma2-9B} appear more targeted (e.g., Medicine), while \texttt{RoLlama3.1-8B} shows less consistent visual improvement over its base.
\end{itemize}

Overall, the heatmap confirms that performance varies significantly by topic, influenced by model scale and targeted fine-tuning.

\subsection{Performance by Difficulty Levels}

Table~\ref{tab:difficulty_comparison} summarizes accuracy based on question difficulty.

\begin{table}[ht]
\resizebox{0.9\columnwidth}{!}{
\begin{tabular}{lccc}
\toprule
\textbf{Model} & Easy & Medium & Hard \\
\midrule
RoGemma2-9B & 84.7 & 79.2 & 84.2 \\
Gemma2-9B & 84.3 & 77.9 & 73.6 \\
Qwen2.5-72B & 86.4 & 82.8 & \textbf{94.7} \\
Llama-3.3-70B & \textbf{91.4} & \textbf{86.4} & 89.4 \\
\bottomrule
\end{tabular}}
\caption{Accuracy (\%) by question difficulty level.}
\label{tab:difficulty_comparison}
\end{table}

\paragraph{Observations:}
\begin{itemize}
    \item \texttt{Llama-3.3-70B} and \texttt{Qwen2.5-72B} show robustness across all difficulty levels.
    \item Smaller models generally see significant accuracy drops on medium and hard questions.
    \item Romanian-fine-tuned models often outperform their base models at higher difficulties.
\end{itemize}

\subsection{Cross-Language Testing on French and English Datasets}
% \todo{de adaugat rezultate si comentarii}

To investigate whether observed performance patterns generalize beyond Romanian, we evaluated models on comparable French and English datasets. For each language, we used a sample consisting of 35 hard, 165 medium, and 300 easy questions. Table~\ref{tab:language_comparison} shows the comparative accuracy.

\begin{table}[ht]
\resizebox{0.9\columnwidth}{!}{
\begin{tabular}{lccc}
\toprule
\textbf{Model} & Romanian & French & English \\
\midrule
Gemma2-9B & 81.4  & 76.4 & 88.8 \\
Llama-3.1-8B & 71.8 & 68.8 & 85.0 \\
Aya-23-8B & 73.4 & 66.0 & 77.6 \\
Mistral-7B & 52.6 & 47.2 & 75.6 \\
EuroLLM-9B & 74.2 & 70.6 & 79.6 \\
Qwen2.5-72B & 68.2 & 83.1 & 93.6 \\
Llama-3.3-70B & 77.0 & 82.2 & 94.0 \\
\bottomrule
\end{tabular}}
\caption{Accuracy (\%) by language.}
\label{tab:language_comparison}
\end{table}

A consistent performance hierarchy emerges across models, with accuracy highest in English, followed by Romanian, and lowest in French (Table~\ref{tab:language_comparison}). This likely reflects the dominance of English in pre-training data. The lower French scores compared to Romanian suggest language-specific challenges or dataset differences beyond language family. While larger models like \texttt{Qwen2.5-72B} and \texttt{Llama-3.3-70B} show greater cross-lingual robustness, consistent performance across languages remains challenging—highlighting the need for continued language-specific evaluation in multilingual NLP.

\subsection{Translation-Based Comparison with English}
% \todo{de adaugat rezultate si comentarii}

% In order to isolate the impact of linguistic versus cultural grounding, we translated all Romanian questions into English and re-evaluated model performance on these translations. By comparing responses to the original Romanian prompts and their English equivalents, we could analyze whether discrepancies stemmed from model limitations in understanding Romanian semantics or from a lack of alignment with culturally specific content.
To isolate the impact of linguistic versus cultural grounding, we translated all Romanian questions into English and re-evaluated model performance. By comparing responses to the original and translated versions, we analyzed whether discrepancies arose from limitations in Romanian language understanding or from gaps in cultural knowledge.

%\begin{table}[ht]
%\resizebox{0.9\columnwidth}{!}{
%\begin{tabular}{lcc}
%\toprule
%\textbf{Model} & Original & Translated \\
%\midrule
%Gemma2-9B & 81.8 & 80.2 \\
%EuroLLM-9B & 83.3 & 76.3 \\
%Qwen2.5-72B & 85.9 & 83.3 \\
%Llama-3.3-70B & 90.7 & 85.0 \\
%\bottomrule
%\end{tabular}}
%\caption{Comparison between original and translated Romanian dataset.}
%\label{tab:original_vs_translated}
%\end{table}
\begin{figure}[ht]
    \centering
    \includegraphics[width=0.9\linewidth]{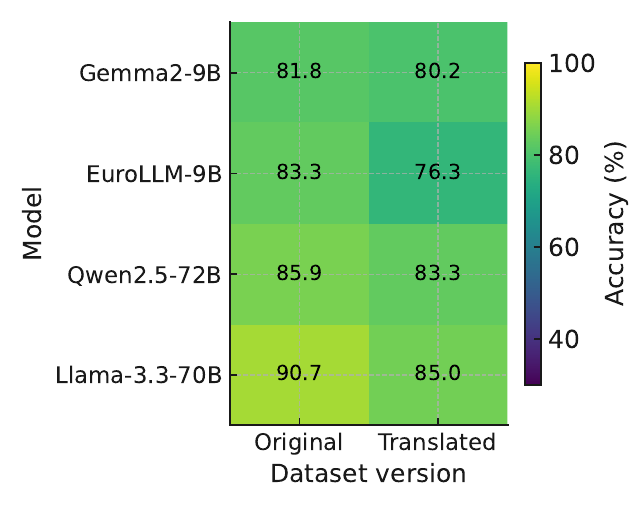}
    \caption{Comparison between original and translated Romanian dataset.}
    \label{fig:original_vs_translated}
\end{figure}

This experiment revealed that translation did not help and, in fact, led to slightly worse performance, highlighting the importance of native-language benchmarks for fair and accurate evaluation.

\subsection{Illustrative Examples of Culturally-Specific Questions}
\label{sec:illustrative_examples}

To illustrate the nature of the challenges posed by the Romanian-specific subset (discussed in Section~\ref{sec:cultural_context_results}), here are selected examples requiring localized knowledge. English translations are provided.

\begin{itemize}
   \item \textbf{Historical Context (Index 896):} 
    \textit{RO:} În afară de Marea Unire, domnia lui Ferdinand I a fost marcată și de: 
    (\textit{EN:} Besides the Great Union, the reign of Ferdinand I was marked by:) 
    Answer: \textit{Primul Război Mondial / World War I}. 
    Model prediction: \textit{loss of Bucovina}. 
    \textit{Commentary:} The English translation misses the Romanian historical context, where Ferdinand’s reign is more closely associated with WWI than the loss of Bucovina.
    
    \item \textbf{Geographical Knowledge (Index 958):} 
    \textit{RO:} Ce munți trebuie să urci ca să vizitezi Babele? 
    (\textit{EN:} What mountains must you climb to visit Babele?) 
    Answer: \textit{Bucegi}. 
    Model prediction: \textit{Făgăraș}. 
    \textit{Commentary:} The English translation doesn't account for the specific location of Babele in the Bucegi Mountains, leading to the incorrect prediction of Făgăraș.

    \item \textbf{Idiomatic Expression (Index 918):} 
    \textit{RO:} O expresie veche românească spune că omul care speră lucruri irealizabile visează: 
    (\textit{EN:} An old Romanian expression says that the man who hopes for unachievable things dreams:)  
    Answer: \textit{cai verzi pe pereți / green horses on walls}.  
    Model prediction: \textit{green and dried}.  
    \textit{Commentary:} A literal translation misses the idiomatic meaning, which is culturally rooted and nonsensical without Romanian-specific knowledge.
    
\end{itemize}
These questions exemplify how the dataset probes knowledge beyond internationally common facts, demanding culturally embedded understanding.

\section{Conclusions}
% \todo{Continua concluziile dupa experimentele cu franceza si engleza}

We introduced \textbf{WWTBM}, a novel, culturally-rich multilingual dataset derived from "Who Wants to Be a Millionaire?" videos, designed for evaluating LLMs on Romanian cultural nuances. Our benchmarking revealed several key insights into model performance at the intersection of language and culture.

A \textbf{significant performance gap} consistently appears between international (high accuracy) and \textbf{Romanian-specific questions} (lower accuracy) across all tested models, including large-scale ones like \texttt{Llama-3.3-70B-Instruct}. This highlights that current pre-training does not fully capture deep cultural knowledge. Furthermore, Romanian-specific fine-tuning showed \textit{mixed effects}, sometimes improving linguistic adaptation more than cultural knowledge retrieval.

Cross-lingual tests confirmed performance variations across languages (English > Romanian > French). Notably, translating Romanian questions to English \textbf{decreased accuracy} compared to the original, underscoring the value and potential subtleties of \textbf{native-language benchmarks}.

Overall, our findings stress the critical need for \textbf{culturally-grounded datasets} like WWTBM for robust LLM evaluation. Developing models adept across diverse cultural contexts remains challenging but crucial, particularly for building effective and equitable \textit{educational NLP applications}. WWTBM provides a valuable resource for advancing this research.

\section{Ethical Considerations}

The dataset used in this research was obtained solely from publicly available video recordings of the Romanian edition of \textit{Who Wants to Be a Millionaire?}, accessible on platforms such as YouTube. No private or restricted content was involved.

The dataset is released as an open-source resource to support multilingual and culturally-aware NLP research. No personally identifiable information (PII) of participants or audience members was collected, included, or distributed. The research team ensured that all content is educationally relevant and intended strictly for research and development purposes.

We recognize the importance of ongoing ethical oversight and remain committed to addressing any concerns raised by content owners or stakeholders. Future users are encouraged to uphold these standards and use the dataset responsibly and in accordance with applicable copyright.

\section{Limitations}

While this study provides valuable insights into multilingual LLM performance in Romanian-specific and international contexts, several limitations must be acknowledged.

First, the dataset is relatively small, with around 1,000 questions collected from available recordings of the Romanian edition of \textit{Who Wants to Be a Millionaire?}. Despite careful curation, this limited size may affect the statistical robustness of evaluations, particularly for difficult questions and less common topics. Moreover, the dataset reflects the knowledge areas, difficulty distribution, and cultural biases of the quiz show format, which may not fully represent broader Romanian culture or general knowledge domains.

Second, the use of automated tools—such as multimodal OCR and LLM-based categorization—may introduce annotation errors, even with subsequent manual correction. While manual validation helped reduce inconsistencies, some residual inaccuracies may have influenced results.

Third, our evaluation was limited to multiple-choice questions with minimal context, which may oversimplify real-world language understanding. Results could differ in tasks involving open-ended answers, reasoning, or richer context.

Lastly, the study focused on open-source LLMs. Evaluations including proprietary or domain-specialized models might produce different outcomes.

Future work should expand the dataset in size and diversity, include varied question formats, and explore tasks requiring deeper reasoning and contextual understanding to better guide Romanian-language and educational NLP development.

\bibliography{references}

% \todo{Mai verificati o data referintele, le am verificat dar trebuie double checked, sunt multe si mi-e ca pe unele nu le-am scris corect, pe unele nu cred ca le am completat, trebuie verificat si daca jurnmalul este s
% cris corect}

\appendix % Corrected: removed stray 'l'. Standard placement is often before \bibliography.

\section{Few-Shot Learning Results}
\label{sec:appendix_few_shot}

To further investigate the impact of Romanian fine-tuning and provide additional context beyond the zero-shot evaluations presented in the main paper, we conducted experiments using few-shot prompting. We evaluated model performance by providing 1, 3, or 5 examples (shots) from the training portion of our dataset within the prompt before presenting the actual test question. The examples were randomly selected for each query.

Table~\ref{tab:few_shot_results_single_col} presents the accuracy scores for various base models and their corresponding Romanian fine-tuned variants across 0, 1, 3, and 5-shot settings. Due to space constraints, model names are abbreviated (e.g., RLa2=RoLlama2, M3.1=Llama-3.1, RGe2=RoGemma2, DPO dates omitted where ambiguous).

\begin{table}[ht!]
\centering
\small % Make font smaller
\setlength{\tabcolsep}{3pt} % Reduce space between columns
\begin{tabular}{lccccc}
\toprule
\textbf{Model Abbrev.} & \textbf{FS=0} & \textbf{FS=1} & \textbf{FS=3} & \textbf{FS=5} \\
& (\%) & (\%) & (\%) & (\%) \\
\midrule
Llama-2-7b              & 29.7 & 48.2 & 52.5 & 52.1 \\
RLa2-7b-Base            & 25.0 & 35.0 & 46.4 & 50.7 \\
\midrule
Llama-2-7b-Chat         & 48.3 & 52.4 & 56.4 & 55.7 \\
RLa2-7b-Ins (Oct24)     & 68.6 & 72.6 & 75.5 & 74.0 \\
RLa2-7b-Ins (Apr25)     & 63.7 & 73.2 & 75.3 & 74.0 \\
RLa2-7b-DPO (Oct24)     & 59.7 & 69.2 & 74.4 & 72.8 \\
RLa2-7b-DPO (Apr25)     & 71.7 & 71.6 & 73.4 & 73.2 \\
\midrule
Mistral-7B-Ins-v0.3   & 67.0 & 67.3 & 69.7 & 69.6 \\
RMi-7b-Ins (Oct24)      & 68.7 & 71.7 & 75.6 & 76.1 \\
RMi-7b-Ins (Apr25)      & 73.5 & 74.1 & 74.5 & 76.1 \\
RMi-7b-DPO (Oct24)      & 48.5 & 70.8 & 73.2 & 74.0 \\
RMi-7b-DPO (Apr25)      & 73.6 & 73.1 & 73.9 & 74.4 \\
\midrule
Llama-3-8B-Ins          & 67.9 & 71.5 & 71.3 & 71.6 \\
RLa3-8b-Ins (Oct24)     & 72.8 & 75.3 & 75.9 & 75.8 \\
RLa3-8b-Ins (Apr25)     & 74.3 & 75.6 & 77.4 & 76.1 \\
RLa3-8b-DPO (Oct24)     & 73.8 & 75.3 & 75.3 & 75.9 \\
RLa3-8b-DPO (Apr25)     & 76.5 & 75.6 & 76.8 & 77.1 \\
\midrule
Llama-3.1-8B-Ins      & 70.1 & 71.2 & 72.1 & 72.8 \\
RLa3.1-8b-Ins (Oct24)   & 72.6 & 74.6 & 76.0 & 77.5 \\
RLa3.1-8b-Ins (Apr25)   & 73.1 & 75.1 & 76.7 & 76.6 \\
RLa3.1-8b-DPO (Oct24)   & 74.1 & 74.0 & 74.9 & 76.6 \\
RLa3.1-8b-DPO (Apr25)   & 76.1 & 74.8 & 74.5 & 75.2 \\
\midrule
Gemma-1.1-7b-IT         & 65.1 & 61.9 & 61.1 & 60.4 \\
RGe-7b-Ins (Oct24)      & 71.7 & 74.4 & 75.5 & 76.2 \\
RGe-7b-Ins (Apr25)      & 70.6 & 76.1 & 76.0 & 76.6 \\
RGe-7b-DPO (Oct24)      & 70.6 & 74.2 & 75.1 & 75.8 \\
\midrule
Gemma-2-9b-IT           & 79.6 & 81.1 & 81.9 & 82.5 \\
RGe2-9b-Ins (Oct24)     & 78.4 & 80.9 & 82.3 & 81.1 \\
RGe2-9b-Ins (Apr25)     & 75.8 & 74.9 & 79.3 & 79.5 \\
RGe2-9b-DPO (Oct24)     & 79.5 & 80.5 & 82.2 & 81.1 \\
RGe2-9b-DPO (Apr25)     & 81.4 & 78.9 & 80.5 & 80.4 \\
\bottomrule
\end{tabular}
\caption{Few-Shot (FS) Accuracy Comparison: Base Models vs. Romanian Fine-tuned Variants. Model names abbreviated for space.}
\label{tab:few_shot_results_single_col}
\end{table}

\paragraph{Brief Observations:}
These few-shot results generally confirm the zero-shot trends. Romanian fine-tuned models often maintain or slightly increase their advantage over base models with 1-5 examples. Performance tends to saturate or slightly decrease between 3 and 5 shots. DPO variants are competitive, sometimes outperforming Instruct versions, especially at 0/1 shot. Base Gemma models show less benefit from few-shot examples here. A detailed analysis is future work.

\end{document}